# Rule Extraction Algorithm for Deep Neural Networks: *A Review*

Tameru Hailesilassie

Department of Computer Science and Engineering
National University of Science and Technology (MISiS)
Moscow, Russia
tameruh.s@gmail.com

*Abstract*—Despite the highest classification accuracy in wide varieties of application areas, artificial neural network has one disadvantage. The way this Network comes to a decision is not easily comprehensible. The lack of explanation ability reduces the acceptability of neural network in data mining and decision system. This drawback is the reason why researchers have proposed many rule extraction algorithms to solve the problem. Recently, Deep Neural Network (DNN) is achieving a profound result over the standard neural network for classification and recognition problems. It is a hot machine learning area proven both useful and innovative. This paper has thoroughly reviewed various rule extraction algorithms, considering the classification scheme: decompositional, pedagogical, and eclectics. It also presents the evaluation of these algorithms based on the neural network structure with which the algorithm is intended to work. The main contribution of this review is to show that there is a limited study of rule extraction algorithm from DNN.

## I. INTRODUCTION

The dramatic advances in technology nowadays require defined computing features that are not available on Von Neumann computers. Those characteristics include learning ability, generalization ability, and adaptivity [1]. Researchers are doing much work to meet this demand by mimicking biological neural system to model a modern computer system which is much closer to a human nervous system. Digital computers surpass human in solving computational problems. However, a person can solve complex perceptual problems. Accordingly, understanding the way the brain solves such problems benefit computing system enormously. Table I shows the comparison of a digital computer and the human brain.

Artificial neural networks are the inspiration of human nervous system formed by an interconnection of neurons, a single processing unit. Thousands and ten thousands of transistors in a computer are analogous to neurons in the neural network. Nevertheless, neural networks have a significant number of connections, whereas computers have few [2]. The ability of high noise tolerance, adaptive learning, fault tolerance, a highly accurate classification for the big dataset, and self-organization makes a neural network fit easily in a range of application areas. Application areas include, but not limited to, industrial process control, sensory data recognition, medical diagnosis, sales forecasting, customer research, pattern recognition, and so on.

In the case of recognition, to achieve excellent performance, we need a large dataset to train a neural network. However, training requires a model with high learning ability and deep processing capacity. Research has shown that deep processing also takes place in the human brain [3]. Consequently, there is a need for deep processing. Deep neural network (DNN) is intended for this need. Feedforward network has at most one or two hidden layers, but DNN has a stack of more than two hidden layers as depicted in Figure 1. The deep depth, number of hidden layers, in DNN is useful to work with complex problems more efficiently than the shallow artificial neural network. Recently, DNN has drawn researchers' attention. Especially, it becomes dominant for video, audio, and Image classification and retrieval [4], [5], [6]. As well as, the significant result has been achieved for object detection [7], [8].

TABLE I. COMPARISON OF STANDARD COMPUTER (CIRCA 1994) AND HUMAN BRAIN [9]

|  | **Computer** | **Human Brain** |
|---|---|---|
| Computational units | 1 CPU, $10^5$ Gates | $10^{11}$ Neurons |
| Storage units | $10^9$ bits RAM, $10^{10}$ bits disk | $10^{11}$ Neurons, $10^{14}$ Synapses |
| Cycle time | $10^{-8}$ Sec | $10^{-3}$ sec |
| Bandwidth | $10^9$ bits/sec | $10^{14}$ bits/Sec |
| Neuron updates/sec | $10^5$ | $10^{14}$ |

Notwithstanding neural network attains high classification performance, they are not easily understandable [10]. All the promising results achieved by the neural networks are within a black box approach which is not comprehensible by the human. The shallow artificial neural network and DNN have many applications in safety and mission critical systems. Some of theme are: industrial process control and fault detection, power generation and transmission, aircraft icing ( weather forecasting to assist pilots), consumer products such as LogiCook (the first neural network micro oven), medical diagnosis, vehicle health monitoring, and many others [11]. However, safety-critical systems require reliability validation, otherwise it leads to danger [12]. Hence, to change neural network black box system into white box system, rule extraction techniques from artificial neural network are studied in depth for the previous two decades. Rule extraction is an approach to reveal the hidden knowledge of the network and help to explain the process how neural network comes to a final decision. So that user can understand it better. Moreover, extracted rules can be used for hazard mitigation, traceability, system stability and fault tolerance, operational verification and validation, and more [13].

Deep learning is showing astonishing result lately. Researchers are reporting that DNN outperforms over the standard artificial neural network in several areas of Machine leraning. Additionaly, extracting a comprehensible rule from DNN enhance the power and acceptability of DNN products, that comprises both comprehensibility and accuracy. Many rule extraction algorithms have proposed by researchers. Our review focuses on rule extraction algorithms from the neural network. In this work, we do not discuss algorithms that utilize neural network for rule extraction as a tool.

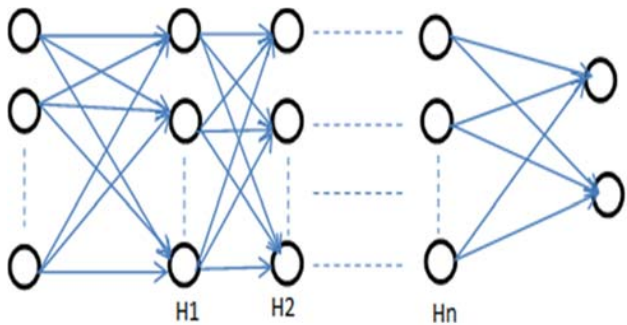

Figure 1. Sample DNN architecture

The aim of this paper is:

A. *to discuss various rule extraction techniques from neural network*

B. *to analyze the applicability of the existing rule extraction algorithms for DNN*

C. *to address challenges of rule extraction from DNN and to provide future research direction*

## II. RULE EXTRACTION ALGORITHM

According to [14] rule extraction is defined as "…given a trained neural network and the data on which it was trained, produce a description of the network's hypothesis that is comprehensible yet closely approximates the network's predictive behavior." Rule extraction algorithm is useful for experts to verify and cross-check neural network systems. Extracted rules have different forms. We present an overview of some of the logical rules below.

**IF-THEN rules**: It is a conditional statement model, which is easily comprehensible. The general form of IF-THEN rule is:

$$IF\ X \epsilon S(i)\ THEN\ Y=y(i) \quad (1)$$

If the given condition is true, in this case if X is a member of $S(i)$ then the output will be labeled to a particular class. As a simple example, a single neuron in a neural network with a linear activation function can be modeled by IF-THEN logical rule. The weighted sum of $j^{th}$ neuron is calculated as:

$$S_j = \sum_i^n (X_i * W_{ij}) \quad (2)$$

Where, $X_i$ is an input and $W_{ij}$ is a corresponding weight of connection between $i^{th}$ and $j^{th}$ neuron. The output Y of a neuron is a function of the weighted sum given as:

$$Y=f(S_j) \quad (3)$$

So that, the output can be expressed by simple IF-THEN rule as follow:

$$IF\ Y=f(S_j) \geq \alpha\ THEN\ Y=1\ ELSE\ Y=0,$$

Where α is a Threshold value.

**M-of-N rules:** It searches for rules with a Boolean expression. The expression is satisfied when M of N sets are satisfied. The rule has the following form:

$$IF\ M\ of\ \{N\}\ THEN\ Z$$

This method is efficient and general [15]. It can also easily converted to a simple IF-THEN rule.

**Decision tree:** It is a most widely used tree structure classifier in Machine learning and Data mining. This model classifies an instance starting at the root of the tree and follow down to the branches till the end. Decision tree uses a white box system that is easy to explain. A simple Decision tree diagram is shown in Figure 2.

Andrew, Diederich, and Tickle proposed a multidimensional taxonomy for the various rule extraction algorithm [12]. The first dimension of classification is based on the expressive power of the extracted algorithm, which refers to the form of the rule (e.g. IF-THEN rule and Fuzzy rule). The second scheme considers the relationship between the extracted rule and the architecture of trained neural network. Accordingly, there are three categories. Rule extraction algorithms that work on Neuron- level, rather than the whole neural network architecture-level are called decompositional techniques. If artificial neural network is considered as a black box irrespective of the architecture, then these algorithms fall into a pedagogical category. The third is the combination of both decompositional and pedagogical approaches. It is called eclectics.

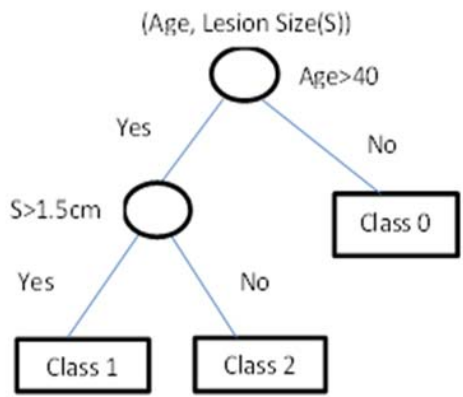

Figure 2. Sample Decision Tree

### A. DECOMPOSITIONAL APPROACH

Decomposition algorithms work by splitting the network into neuron level. The result obtained from each neuron then aggregated to represent the network as a whole. Özbakır, Baykasoğlu, and Kulluk [16] have introduced an algorithm called DIFACON-miner that can generate an IF-THEN rule from artificial neural network. Differential evolution (DE) and touring ant colonization (TACO) algorithm are used for training and rule extraction respectively. The rule generation takes place in each iteration of DE. Before this proposed algorithm, rule extraction was a sequential process. What makes this work different is that neural network training and rule generation can be performed simultaneously. It saves additional quite a long time spent for rule extraction. However, the algorithm addresses only a feedforward artificial neural network. This algorithm does not consider rule extraction from DNN. An algorithm called CRED [17] extract both continuous and discrete rules by using decision tree from pre-trained ANN. Algorithms that specifically work with discrete inputs requires discretization of continuous attributes to comprise continues data which in turn affects the accuracy of extracted rule. Thus, CRED does not employ discretization. Also, the authors have proposed an algorithm for rule simplification based on J-measure. This algorithm showed a good result tested on UCI database. Even though this algorithm is independent of network structure and training algorithm, it is not possible to apply the algorithm directly to DNN. FERNN [18] is an algorithm proposed to generate both M-of-N and IF-THEN rule from feedforward neural network with a single hidden layer. Besides, there is no pruning and retraining of the network. As a result, the speed of extraction process is fast. The applicability of the algorithm to DNN is not even mentioned. Fu has proposed an algorithm called KT [19] that extract IF-THEN rule. It follows layer by layer approach. This algorithm applies a tree search to find the rule. In spite of the fact that the extracted rules are robust, in some cases even better than Neural network itself, DNN is not taken into consideration. Tsukimoto [20] has introduced a rule extraction algorithm that works with different types of neural network, such as recurrent neural network and multilayer perceptron (MLP), whose activation function is monotonic. It extracts IF-THEN rules that are applicable for both continuous and discrete values. Also, the computational complexity of this algorithm is polynomial, whereas KT is in the order of exponential.

### B. PEDAGOGICAL APPROACH

Neural network is treated as a black box system in pedagogical rule extraction algorithms. The focus of this approach is finding an output for a corresponding input. The weight of internal structure of artificial neural network is not subjected to analysis [21]. There are different techniques in this approach. Some of them are validity interval analysis (VIA) [22], sampling approach, and reverse engineering the neural network.

Craven and Shavlik [23] have proposed a pedagogical algorithm called TREPAN. It extracts M-of-N split point and decision tree from ANN by utilizing query and sampling approach. Learning with queries is also introduced with the aim of information retrieval, which is essential for the learning process. The architecture of neural network used for experiment has only one hidden layer. Saad and Wunsch [24] have introduced a pedagogical approach method HYPINV, based on network inversion technique. This algorithm is capable of extracting hyperplane rule. The rule extraction is in the form of conjunction and disjunction of hyperplanes. The authors used a standard MLP for the experiment. Notwithstanding that the algorithm is independent of the architecture of MLP, DNN is not contemplated. Taha and Ghosh [25] have proposed three rule extraction algorithms. The first one is a pedagogical approach named as BIO-RE. It extracts a binary rule from a neural network which is trained with binary input. However, the algorithm is tested with a shallow MLP of four input, six hidden and three output node.

*(IJCSIS) International Journal of Computer Science and Information Security,*
*Vol. 14, No. 7, July 2016*Sethi, D. Mishra, and B. Mishra [26] have proposed KDRuleEx that can generate a two-dimensional matrix of a decision table. Training example set and trained artificial neural network are input for this algorithm. It can work with both discrete and continuous inputs. Also, unlike most other algorithms, it can handle non-binary inputs. The authors did not report the architecture of artificial neural network used for the test. Moreover, DNN is not mentioned at all. A reverse engineering approach algorithm called RxREN which is proposed by Augusta and Kathirvalavakumar [27] extracts an IF-THEN rule from a neural network. Reverse engineering techniques are the analysis of output, to trace back components that cause the final result. The authors reported that this algorithm is fast to search the rules. Furthermore, only a conventional feedforward neural network is used for an experiment. Schmitz, Aldrich, and Gouws [28] have proposed ANN-DT. This algorithm uses the sampling technique. It extracts a binary decision tree from a feedforward neural network with both discrete and continuous data.

In contrast with pedagogical algorithms, a decompositional approach is much translucent. However, the decompositional technique works layer by layer. As a result, this method may be tedious and time-consuming. Regarding computational limitation and execution time, pedagogical approach is better than decompositional [29]. Also, it has an advantage of flexibility in terms of ANN architecture.

*C. ECLECTIC APPROACH*

Hruschka and Ebecken [30] have proposed an algorithm which is based on an algorithm called RX proposed by Lu, Setiono, and Liu [31]. The technique incorporates both decompositional and pedagogical approach. Clustering genetic algorithm (CGA) is employed for the purpose of hidden unit clustering. Subsequently, a logical rule extraction is based on the relationship between input and generated cluster. This algorithm is designed for shallow MLP neural network. An eclectic approach algorithm that uses artificial immune system (AIS) is proposed by Kahramanli and Allahverdi [32]. It can generate a rule from a trained shallow feed forward neural network. Also, they reported that it has high accuracy value.

*D. RULE EXTRACTION FROM DEEP NEURAL NETWORK*

Zilke [33] have proposed a rule extraction algorithm called DeepRED from DNN by extending a decompositional algorithm CRED. The proposed algorithm has additional decision trees as well as intermediate rules for every hidden layer. Rule extraction from DNN is a step-wise process. It is a divide and conquer method that describes each layer by the previous layer. Accordingly, each result is merged to get the final rule that explains the whole DNN. The author has found the divide and conquer approach helpful to reduce memory usage and computational time. Moreover, to facilitate the process of rule extraction, FERNN is used for pruning less important components of a neural network. Also, it is reported that the experimental test with different datasets showed a high level of accuracy. The author said it might be the first rule extraction algorithm from DNN.

III. FUTURE DIRECTION

The review presented in the previous section clearly demonstrates that there have been a lot of works devoted to rule extraction from artificial neural network. Table II shows the summary of rule extraction algorithms along with the form of extracted rule and used neural network type for the experiment.

TABLE II. SUMMARY OF ALGORITHMS

| Algorithm | Used ANN type | Algorithm Type | Extracted Rule form |
|---|---|---|---|
| DIFACON-miner | Standard MLP | Decompositional | IF-THEN |
| CRED | Standard MLP | Decompositional | Decision tree |
| FERNN | Standard MLP | Decompositional | M-of-N, IF-THEN |
| KT | Standard MLP | Decompositional | IF-THEN |
| Tsukimoto's Algorithm | Standard MLP and RNN | Decompositional | IF-THEN |
| TREPAN | Standard MLP | Pedagogical | M-of-N spilit, decision tree |
| HYPINV | Standard MLP | Pedagogical | Hyperplane rule |
| BIO-RE | Standard MLP | Pedagogical | Binary rule |
| KDRuleEX | Standard MLP | Pedagogical | Decision tree |
| RxREN | Standard MLP | Pedagogical | IF-THEN |
| ANN-DT | Standard MLP | Pedagogical | Binary Decision tree |
| RX | Standard MLP | Eclectic | IF-THEN |
| Kahramanli and Allahverdi's Algorithm | Standard MLP | Eclectic | IF-THEN |
| DeepRED | DNN | Decompositional | IF-THEN |

From Table II, we can understand that the target of most rule extraction algorithms is standard MLP neural network. So far there is a limited study concerning DNN. The challenge of DNN is the complexity of its hidden layers. However, one could develop an algorithm to extract a comprehensible rule from DNN by taking advantage of Pedagogical approach. Unlike decompositional approach, a pedagogical technique is not affected by the number of hidden layers. The recent astonishing ability of DNN to solve a variety of complex problems can be further improved by extracting understandable rules. Consequently, we can use DNN for different problem domains where validation is essential. This paper argues that rule extraction from DNN is important to take advantage of the performance. Besides, the existing work is not adequate. Therefore, this area still requires particular attention.

IV. CONCLUSION

This paper attempts to provide a review of several rule extraction algorithms from an artificial neural network. Some of the state-of-the-art algorithms are discussed from each category named as Decompositional, Pedagogical, and Eclectics. Currently, Deep Learning provides an acceptable solution for lots of problems. It is a new machine learning area which is believed to move machine learning a step ahead. The review implies that, surprisingly, little work has done targeting DNN. It is still a black box system. Even though DNN architecture is complex, a pedagogical algorithm can be used as an advantage irrespective of the number of hidden layer. Pedagogical algorithms do not depend on the architecture of algorithm. Thus, they might fill this gap. Extracting a comprehensible rule from DNN enhance the real world usability of the promising solutions of DNN. Also, it can remove uncertainty problems associated with neural network software.

REFERENCES

[1] A. K. Jain, J. Mao, and K. M. Mohiuddin, "Artificial neural networks: A tutorial," *Computer*, vol. 29, no. 3, pp. 31–44, Mar. 1996.

[2] P. Domingos, *The master algorithm: How the quest for the ultimate learning machine will remake our world*. United States: Basic Civitas Books, 2015.

[3] M. Grégoire, "On layer-wise representations in deep neural networks", Ph.D. dissertation, Technische Universität, Berlin, 2013.

[4] A. Krizhevsky, I. Sutskever and G. E. Hinton, "ImageNet Classification with Deep Convolutional Neural Networks", in *Advances in Neural Information Processing Systems 25*, 2012, pp. 1106--1114.

[5] C. Szegedy, W. Liu, Y. Jia, P. Sermanet, S. Reed, D. Anguelov, D. Erhan, V. Vanhoucke and A. Rabinovich, "Going deeper with convolutions", in *2015 IEEE Conference on Computer Vision and Pattern Recognition (CVPR)*, Boston, MA, 2015, pp. 1 - 9.

[6] H. Lee, P. Pham, Y. Largman and A. Y. Ng, "Unsupervised feature learning for audio classification using convolutional deep belief networks", in *Advances in Neural Information Processing Systems 22 (NIPS 2009)*, 2009, pp. 1096-1104.

[7] C. Szegedy, A. Toshev and D. Erhan, "Deep Neural Networks for Object Detection", in *Advances in Neural Information Processing Systems 26*, 2013, pp. 2553—2561

[8] Szegedy C, Reed S, Erhan D, Anguelov D. Scalable, high-quality object detection. arXiv preprint arXiv:1412.1441. 2014 Dec 3.

[9] S. J. Russell and P. Norvig, *Artificial intelligence: A modern approach*. United Kingdom: Prentice Hall, 1994.

[10] W. Duch, R. Setiono, and J. M. Zurada, "Computational intelligence methods for rule-based data understanding," *Proceedings of the IEEE*, vol. 92, no. 5, pp. 771–805, May 2004.

[11] P. Lisboa, *Industrial use of safety-related artificial neural networks*. HSE Books, 2001.

[12] R. Andrews, J. Diederich and A. Tickle, "Survey and critique of techniques for extracting rules from trained artificial neural networks", *Knowledge-Based Systems*, vol. 8, no. 6, pp. 373-389, 1995.

[13] B. J. Taylor and M. A. Darrah, "Rule extraction as a formal method for the verification and validation of neural networks," *Proceedings. 2005 IEEE International Joint Conference on Neural Networks, 2005*.

[14] M. W. Craven, "Extracting Comprehensible Models from Trained Neural Networks", Ph.D. dissertation, Department of Computer Sciences, University of Wisconsin-Madison, 1996.

[15] G. G. Towell and J. W. Shavlik, "Extracting refined rules from knowledge-based neural networks," *Machine Learning*, vol. 13, no. 1, pp. 71–101, Oct. 1993.

[16] L. Özbakır, A. Baykasoğlu, and S. Kulluk, "A soft computing-based approach for integrated training and rule extraction from artificial neural networks: DIFACONN-miner," *Applied Soft Computing*, vol. 10, no. 1, pp. 304–317, Jan. 2010.

[17] M. Sato and H. Tsukimoto, "Rule extraction from neural networks via decision tree induction", in *International Joint Conference On Neural Network*, Washington, DC, 2001, pp. 1870 - 1875 vol.3.

[18] R. Setiono and W. K. Leow, "FERNN: An algorithm for fast extraction of rules from neural networks," *Applied Intelligence*, vol. 12, no. 1/2, pp. 15–25, 2000.

[19] L. Fu, "Rule generation from neural networks," *IEEE Transactions on Systems, Man, and Cybernetics*, vol. 24, no. 8, pp. 1114–1124, 1994

[20] H. Tsukimoto, "Extracting rules from trained neural networks," *IEEE Transactions on Neural Networks*, vol. 11, no. 2, pp. 377–389, Mar. 2000.

[21] K. KumarSethi, D. Kumar Mishra, and B. Mishra, "Extended Taxonomy of rule extraction techniques and assessment of KDRuleEx," *International Journal of Computer Applications*, vol. 50, no. 21, pp. 25–31, Jul. 2012.

[22] Thrun S. Extracting rules from artificial neural networks with distributed representations. Advances in neural information processing systems. 1995:505-12.

[23] M. Craven and J. Shavlik, "Using Sampling and Queries to Extract Rules from Trained Neural Networks", in *Machine Learning: Proceedings of the 11th International Conference*, San Francisco, CA, 1994.

[24] E. W. Saad and D. C. Wunsch, "Neural network explanation using inversion," *Neural Networks*, vol. 20, no. 1, pp. 78–93, Jan. 2007

[25] I. A. Taha and J. Ghosh, "Symbolic interpretation of artificial neural networks," *IEEE Transactions on Knowledge and Data Engineering*, vol. 11, no. 3, pp. 448–463, 1999

[26] K. Sethi, D. Mishra and B. Mishra, "KDRuleEx: A novel approach for enhancing user Comprehensibility using rule extraction", in *KDRuleEx: A Novel Approach for Enhancing*

*User Comprehensibility Using Rule Extraction*, Kota Kinabalu, 2012, pp. 55 - 60.

[27] M. Augasta and T. Kathirvalavakumar, "Reverse Engineering the Neural Networks for Rule Extraction in Classification Problems", *Neural Process Lett*, vol. 35, no. 2, pp. 131-150, 2011.

[28] G. P. J. Schmitz, C. Aldrich, and F. S. Gouws, "ANN-DT: An algorithm for extraction of decision trees from artificial neural networks," *IEEE Transactions on Neural Networks*, vol. 10, no. 6, pp. 1392–1401, 1999.

[29] M. Augasta and T. Kathirvalavakumar, "Rule extraction from neural networks — A comparative study", in *International Conference on Pattern Recognition, Informatics and Medical Engineering (PRIME-2012)*, Salem, Tamilnadu, 2012, pp. 404 - 408.

[30] E. R. Hruschka and N. F. F. Ebecken, "Extracting rules from multilayer perceptrons in classification problems: A clustering-based approach," *Neurocomputing*, vol. 70, no. 1-3, pp. 384–397, Dec. 2006.

[31] H. Lu, R. Setiono, and H. Liu, "Effective data mining using neural networks," *IEEE Transactions on Knowledge and Data Engineering*, vol. 8, no. 6, pp. 957–961, 1996.

[32] H. Kahramanli and N. Allahverdi, "Rule extraction from trained adaptive neural networks using artificial immune systems," *Expert Systems with Applications*, vol. 36, no. 2, pp. 1513–1522, Mar. 2009.

[33] J. Zilke, "Extracting Rules from Deep Neural Networks", M.S. thesis, Computer Science Department, Technische Universität Darmstadt, 2015.

AUTHORS PROFILE

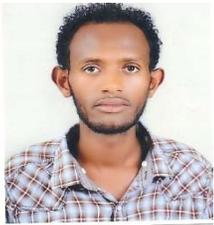

**Tameru Hailesilassie** received a BSc degree in Electrical and Computer Engineering: Computer Engineering focus area from University of Gondar, Institute of Technology with honors in 2015. He is pursuing Master's degree at the Department of Computer Science and Engineering, National University of Science and Technology (MISiS). His research interest includes Software Engineering, Computer Engineering, Machine Learning, and Computer vision.